\documentclass[10pt,twocolumn,letterpaper]{article}

\usepackage{cvpr}
\usepackage{times}
\usepackage{epsfig}
\usepackage{graphicx}
\usepackage{amsmath}
\usepackage{amssymb}
\usepackage{float}

\usepackage[breaklinks=true,bookmarks=false]{hyperref}
\usepackage{cleveref}
\cvprfinalcopy 

\def\cvprPaperID{****} 

\begin{document}

\title{Deep Reinforced Self-Attention Masks for Abstractive Summarization (DR.SAS)}

\author{Ankit Chadha \\
Stanford\\
{\tt\small ankitrc@stanford.edu}
\and
Mohamed Masoud\\
Stanford\\
{\tt\small masoud@stanford.edu}
}

\maketitle

\begin{abstract}

We present a novel architectural scheme to tackle the abstractive summarization problem based on the CNN/DM dataset [1]  which fuses Reinforcement Learning (RL) with UniLM, which is a pre-trained Deep Learning Model, to solve various natural language tasks. We have tested the limits of learning fine-grained attention in Transfomers [3] to improve the summarization quality. UniLM applies attention to the entire token space in a global fashion. We propose DR.SAS which applies the Actor Critic (AC) algorithm [2] to learn a dynamic self-attention distribution over the tokens to reduce redundancy and generate factual and coherent summaries to improve the quality of summarization.  After performing hyperparameter tuning, we achieved better ROUGE results compared to the baseline. Our model tends to be more extractive/factual yet coherent in details because of optimization over ROUGE rewards. We present detailed error analysis with examples of strengths and limitations of our model. Our codebase will be publicly available on our github \cite{DRSAS}

\end{abstract}

\section{Introduction}
\begin{figure*}[htp]
\centering
\includegraphics[height=0.6\textwidth]{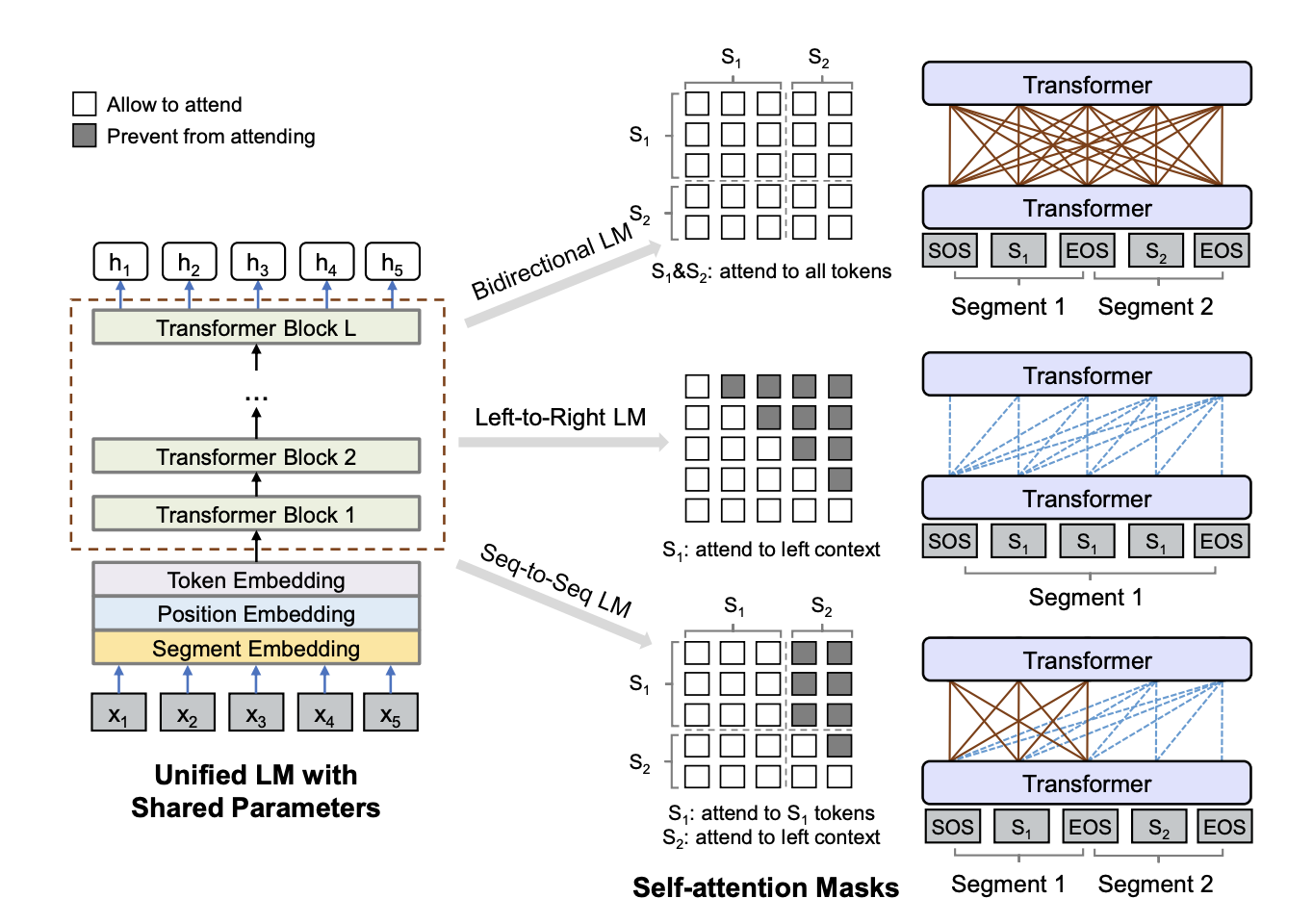}
\caption{UniLM Architecture as presented in \cite{UniLM}}
\label{UNI}
\end{figure*}

\begin{figure*}[htp]
\centering
\includegraphics[scale=0.4]{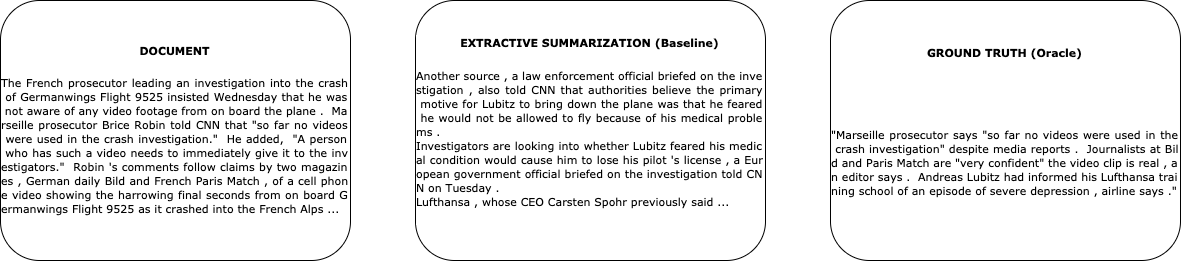}
\caption{(Left to right) Input Document, Baseline(Extractive Summarization), Oracle}
\label{inputoutput}
\end{figure*}

Text summarization is the process of automatically generating natural language summaries from an input document while retaining the important points of the text. This process can be helpful for reducing the cost of decision making by reducing the amount of text that needs to be processed, and can aid many downstream applications such as text completion. Summarization can help peoople with hearing disabilites with applications like voice-to-text and also has use-cases like summarizing medical reports for easy access. Additionally, in this day and age of information explosion, precise summarization can help boost human productivity in daily use cases like news, research and context based search rather than keyword searching \cite{quora}. Generating coherent and meaningful summaries of long text is still a big challenge for most machine learning methods. The abstractive summary, unlike extractive summary, is not constrained to reusing the phrases or sentences in the input text and presents major challenges like saliency, fluency and human readability \cite{Shi}. These challenges stem from the fact that the network tries to optimize an objective that does not involve human quality assessment during training, leading to exposure bias. Models trained using a supervised objective exhibit exposure bias \cite{Paulus}, which is a major challenge for fluency in abstractive models. We are introducing the Actor Critic (AC) reinforcement learning (RL) based technique to reward learning appropriate dynamic self attention masks within the language model, which would improve the model's contextual representations. We modify UniLM \cite{UniLM} which is a SoTA model based on BERT that is pre-trained on multiple downstream language generation tasks (Figure \ref{UNI}). The input to our approach is a document and the output is an abstractive summary of the document (Figure \ref{inputoutput}).

We use the CNN/Daily Mail dataset \cite{Nallapati}, which consists of online news articles and their corresponding summaries. The breakdown of the dataset is: 287226 (92\%) / 13368 (4\%) / 11490 (4\%) train/val/test examples. The dataset is derived from news articles and their corresponding "highlights" as summaries being the ground truth. An example of the dataset is shown in Figure \ref{inputoutput}. We employ ROUGE metrics to evaluate the summaries. ROUGE \cite{Lin} computes a F1 score by comparing matching sub-phrases (n-grams) in the generated summaries against the ground truth \cite{Rouge}. Our approach has two phases. e finetune the environment (UniLM) for the task of abstractive summarization and also employ the Actor-Critic algorithm to learn the per word importance of the abstractive summarization task through policy gradients. Our hypothesis draws  inspiration from \cite{EASummary} where the authors use extractive summarization as input to an abstractive summarization algorithm to generate coherent summaries. However, instead of feeding sentence level priors we will employ Reinforcement Learning (RL) to inherently learn the importance or redundancy of information contributing to coherent summaries.

\section{Related Work}
Researchers have widely utilized recurrent neural encoder-decoder networks to address the abstractive summarization sequence-to-sequence problem \cite{Shi, Nallapati}. These models are unable to represent long documents or generate multiple coherent sentences. To handle the coherence problem, Wu et al \cite{Wu} introduced a neural coherence model which estimates the degree of coherence between two sentences by their distributed representation. They also introduced a Reinforced Neural Extractive Summarization (RNES) model that utilizes RL to add coherence to a neural extractive summarization. 
Lee et al \cite{deepq} estimate Q-values using a neural network that helps maximize ROUGE for Extractive Summarization. The scores are evaluated with the use of word embeddings to capture the similarity between the actual document and the generated summary.
Paulus et al \cite{Paulus} introduced a deep RL method that combines a mixture of supervised sequence-to-sequence modeling to address the abstractive summarization exposure bias introduced by supervised learning. \cite{EASummary}, use a two step approach and employ transformer based models for abstractive summarization. This method uses extractive summarization as a prior to generate abstractive summarization and achieve state-of-the-art (SoTA) ROUGE scores. Generating coherent yet factual summaries has been a challenge for most SoTA approaches where the model traditionally skips over important detail as discussed with examples in \cite{fact}; we aim to address this limitation in abstractive summarization with our approach.
While traditional approaches use predefined self-attention masks or use extractive summarization as a prior to limit redundant information, our approach contains the combined benefits of RL and abstractive summarization to learn the contextual self-attention masks. We hypothesize that this approach will make improvements over the previous methods because the RL objective resembles human quality assessment and abstractive summarization using UniLM generates summaries that are more coherent.

\section{Methods and Architecture}
The presented model studies the effects of using dynamic self-attention masks on the sequence-to-sequence task of abstractive summarization. The model aims to modify the SoTA UniLM and its underlying Transformer architecture to train a reinforcement learning (RL) policy gradient model to generate the self-attention mask. UniLM (UNIfied pre-trained Language Model) contains three types of language modeling tasks: unidirectional, bidirectional, and sequence-to-sequence prediction. UNiLM employs a shared Transformer network and utilizes specific
self-attention masks to control what context the prediction conditions on \cite{UniLM}. Our model specifically tests the use of dynamic self-attention masks on the quality of abstractive summarization. The proposed method trains an Actor Critic (AC) model  [2] to learn a dynamic self-attention distribution over the input tokens with the objective of increasing the summarization output ROUGE score.

\subsection{Actor-Critic Method}
Actor-critic methods are popular deep reinforcement learning algorithms and provide an effective variation of the family of Monte Carlo Policy Gradients (REINFORCE).The policy gradient methods update the probability of stochastic non-deterministic (policy) distribution of actions such that actions with higher expected reward have a higher probability value for an observed state. REINFORCE relies on  estimating policy gradient by sampling using Monte-Carlo methods sample episode and update the policy parameter $\theta$. 

\begin{equation}
\nabla_\theta J(\theta) = \mathbb{E}[\sum_{t=0}^{T-1} \nabla_\theta log (\pi_\theta (a_t|s_t)) Q_{\pi_{\theta}}(s_t,a_t)]
\end{equation}

\begin{equation}
\theta = \theta + \alpha_{\theta} \nabla_\theta J(\theta)
\end{equation}

where $Q_{\pi_{\theta}}(s_t,a_t)$ defines the expected $Q$ value .\\

The Actor-Critic (AC) method provides a variation to the REINFORCE algorithm by defining two modules that formalize the two main components in the policy gradient method: policy model and the value function. 
\begin{enumerate}
    \item The critic computes value functions to help appraise each action the actor takes. It estimates the $Q_{\pi_{\theta}}(s_t,a_t)$ or $V_{\pi_{\theta}}(s_t)$
    \item The actor is the policy that defines how actions are selected and updates the policy parameters $\theta$ for the policy $\pi_\theta (a_t|s_t)$, in the direction suggested by the critic.
    
\end{enumerate}

Basic estimation using policy gradients in  Monte Carlo Policy Learning (REINFORCE) and AC methods is inherently noisy. This could lead to unstable convergence and fluctuations in learning \cite{RLBook}. Reward sparsity can also cause unstable learning \cite{AC1}. The Advantage Actor-Critic (A2C) as described in \cite{A2C} provides a solution to high variance policy distribution by subtracting baseline $b(s_t)$ to reduce variance of the policy gradients. The advantage value $A(s_t,a_t) = Q_{\pi_{\theta}}(s_t,a_t) - b(s_t)$. The advantage actor-critic policy gradient:

\begin{equation}
\nabla_\theta J(\theta) = \mathbb{E}[\sum_{t=0}^{T-1} \nabla_\theta log( \pi_{\theta} (a_t|s_t)) A(s_t,a_t)]
\end{equation}

\begin{figure*}[htp]
\centering
\includegraphics[height=0.6\textwidth]{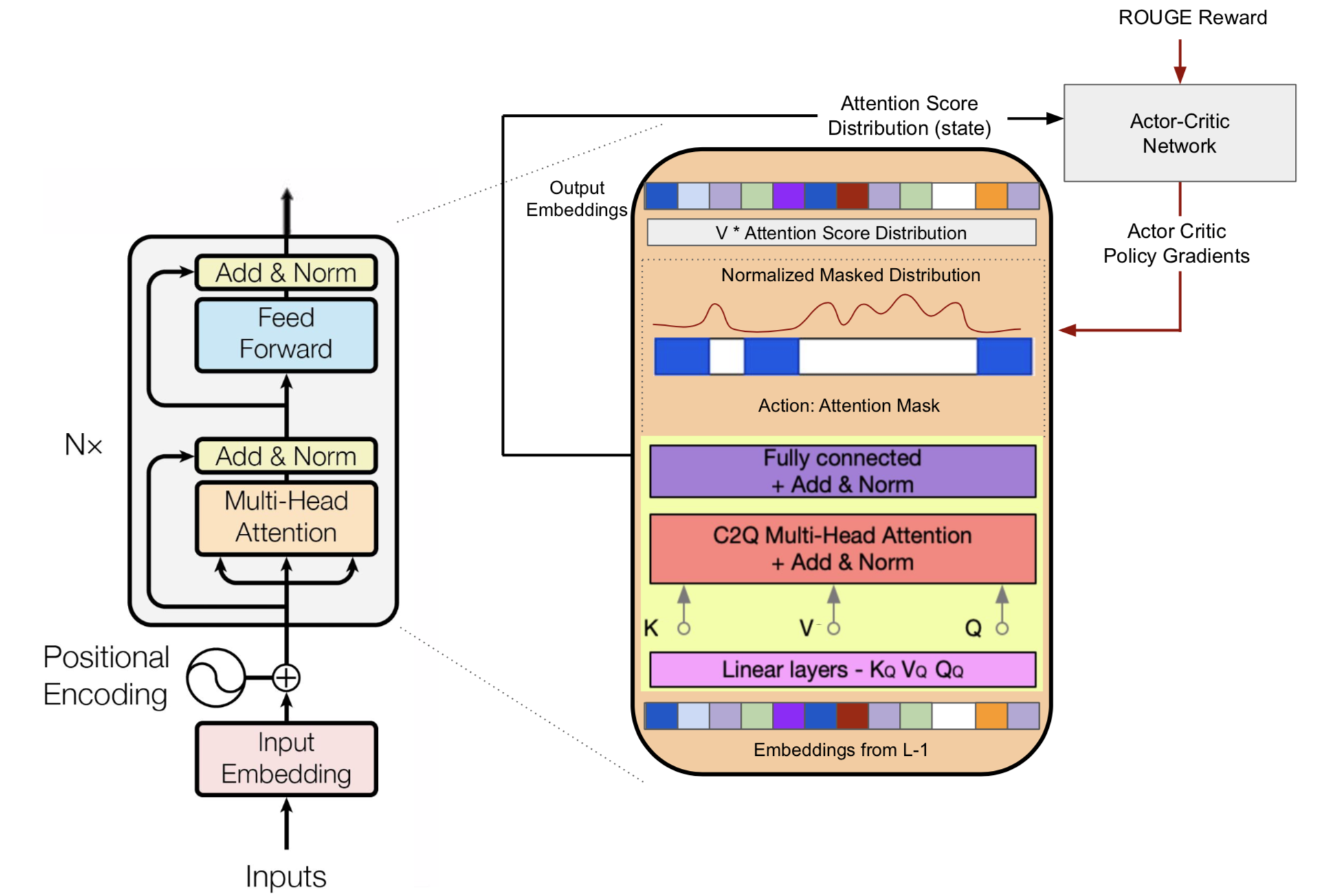}
\caption{Encoder with AC Policy Learning for Self-Attention masks}
\label{txf}
\end{figure*}

\subsection{DR.SAS - Attention Mask Policy Learning (A2C)}

Our model modifies the UniLM underlying Transformer encoder to employ the Advantage Actor-Critic method \cite{Konda} for attention mask policy learning. The A2C framework is to train combined Neural Network and Reinforcement Learning approaches by training the A2C Agent to learn suitable policy based off UniLM + ROUGE environment. The definition of our method with regards to the Actor Critic system is as follows:
\begin{itemize}
\item The state space is described as the attention score distribution from an intermediate layer of UniLM for a given input document to be summarized.

\item The Actor learns a policy distribution and the Critic learns the values $Q_t$. Our method maps this policy to the change in attention masks given the state.

\item The policy distribution defines the probabilities of sampling between two values (0 (mask/not attend), 1 (attend)) ($a_t$) to mask a subset of the input sequence $S$ in \ref{UNI}. The Actor-critic policy gradient will help learn the attention mask for each token given the attention distribution. This essentially should help UniLM ignore redundant information in the document to focus on coherent abstractive summarization.

\item Our model utilizes an environment (UniLM + ROUGE reward function) and uses state-based dynamic attention masks to generate the new summary and produce the ROUGE score and correspondingly learns the updated attention score distribution ($s_{t+1}$).

\item  To reduce variance, we utilize the previous value $V_{\pi_{\theta}}(s_t)$ of the attention score distribution state $s_t$ as the baseline function $b(s_t)$. The updated distribution should not have changed much. 
\end{itemize}  

 The modified encoder with AC Policy Learning for Self-Attention masks is shown in Figure \ref{txf}. We specify in our architecture both the Critic and Actor functions as parameterized neural networks. The Actor-Critic feed forward neural network estimates the value (critic) and the policy (actor) for the continuous action and state spaces. The AC network takes the attention scores (state) of the input tokens and evaluates the state value(s) and policy distribution. The updated policy distribution is then sampled to generate the attention masks. The generated masks are then transformed to values of 0 (attend) and -10000 (mask) to be added to the attention scores that propagate to the UniLM downstream layers to generate the abstractive summary and propagate back the reward (ROUGE) and network gradients. 
 
 The A2C loss is divided into three main components:
 \begin{itemize}
     \item The value (critic) loss:
     \begin{equation}
 \mathcal{L}_{critic} = \frac{1}{2T} \sum_{t=0}^{T-1} A(s_t,a_t) 
\end{equation}
\item The policy (actor) loss:
     \begin{equation}
 \mathcal{L}_{actor} = - \frac{1}{2T} \sum_{t=0}^{T-1} log( \pi_{\theta} (a_t|s_t)) A(s_t,a_t)
\end{equation}

\item The entropy loss which measures the dispersion of the policy distribution $P(x)$:
     \begin{align}
         \mathcal{L}_{entropy} &= \mathbb{E}_{x \sim P}[-log(P_t(x))]\\ &= - \frac{1}{2T} \sum_{t=0}^{T-1} \sum_{i=1}^{N=2} -P_t(x_i) log(P_t(x_i))
     \end{align}

 \end{itemize}
 
The A2C loss is the sum of the three losses:
\begin{equation}
 \mathcal{L}_{a2c} = \mathcal{L}_{actor} + \mathcal{L}_{critic} + \beta \mathcal{L}_{entropy}
\end{equation}

\section{Experimental Results}
In order to evaluate our model, we conducted visual and quantitative analyses. Quantitatively, we evaluate our method in terms of ROUGE. ROUGE stands for Recall-Oriented Understudy for Gisting Evaluation. It is a set of metrics for evaluating automatic summarization of texts as well as machine translation. It works by n-gram comparison between a generated and and reference summary. We compared our model to extractive summarization and the abstractive summarization by UniLM as our baselines. The extractive method selects sentences from a given document to construct a summary of the passage. Specifically, we adapt TextRank \cite{TextRank}, a graph-based ranking model, where the graph vertices are sentences. We use cosine distance with GloVE embeddings \cite{Glove} as a measure of similarity. The baseline can be thought of as a process of recommendation where one sentence recommends another sentence to which it is similar. The baseline was evaluated on the CNN/DM test set. The oracle is essentially a method that can generate the ground truth summaries and would have a ROUGE score of 1. The results in table \ref{results} show a marginal improvement for actor-critic policy gradient additional self-attention mask model compared to TextRank baseline and Fine-Tuned UniLM. The marginal improvement can be attributed to the bias of the reinforced actor-critic policy gradient towards better ROUGE (AC - Reward).

\begin{table}[h]
\centering
\caption{ROUGE Metrics for Baseline, Fine-tuned UniLM and Our Method (Finetuning UniLM in sync with AC)} 
\begin{tabular}{|c|c|c|c|}
\hline
Metrics (F1 score) & Baseline & UniLM & AC + UniLM \\ \hline
ROUGE-1            & 21.0     & 40.79             &   \textbf{41.89}                    \\ \hline
ROUGE-2            & 6.5    & 19.01             &       \textbf{19.22}                \\ \hline
ROUGE-L            & 19.2    & 38.10             &      \textbf{39.28}                \\ \hline
\end{tabular}
\label{results}
\end{table}

\subsection{Hyperparameter Tuning}
Hyperparameter tuning has been a big part for our experiments to achieve generalizability and convergence. A list of hyperparameters we tweaked and tuned for DR.SAS are:
\begin{enumerate}
    \item Max Sequence length - Limited by our local computation resource our initial experiments were done at a sequence length of 384, We eventually switched to 512. This gave us a 0.1 ROUGE score improvement. We attribute this to more context for summarization.

    \item Batch Size - Default: 64, We had to use a batch size of 16 for all our experiments due to compute constraints causing out-of-memory for any batch size beyond. However, we used gradient accumulation in our final experiments to match the default batch size.
    \item Number of epochs - Default: 30, With the limited compute we could only fine-tune for 6 epochs. We compensated for this with a relatively higher learning rate than default.
    \item Learning Rate - Default: 3e-5, We performed random search for the learning rate and settled on 1.5e-4. This gave us relatively faster convergence with fewer epochs.
\end{enumerate}

\subsection{Results Analysis}
 The qualitative analysis show that both UniLM and UniLM + AC produce abstractive summarization of the article by extracting the main points of the article, rewording and connecting the point in coherent way. The UniLM + AC method’s inclination to increase the ROUGE (reward) tends to be to add extractive features in the form of details to the output summary. The n-gram overlap helps increase ROUGE. The bias towards a higher ROUGE score may increase the likelihood that the proposed method extracts less essential points depending on the application. \par
 
\begin{figure*}[htp]
\centering
\includegraphics[width=15cm,height=6cm]{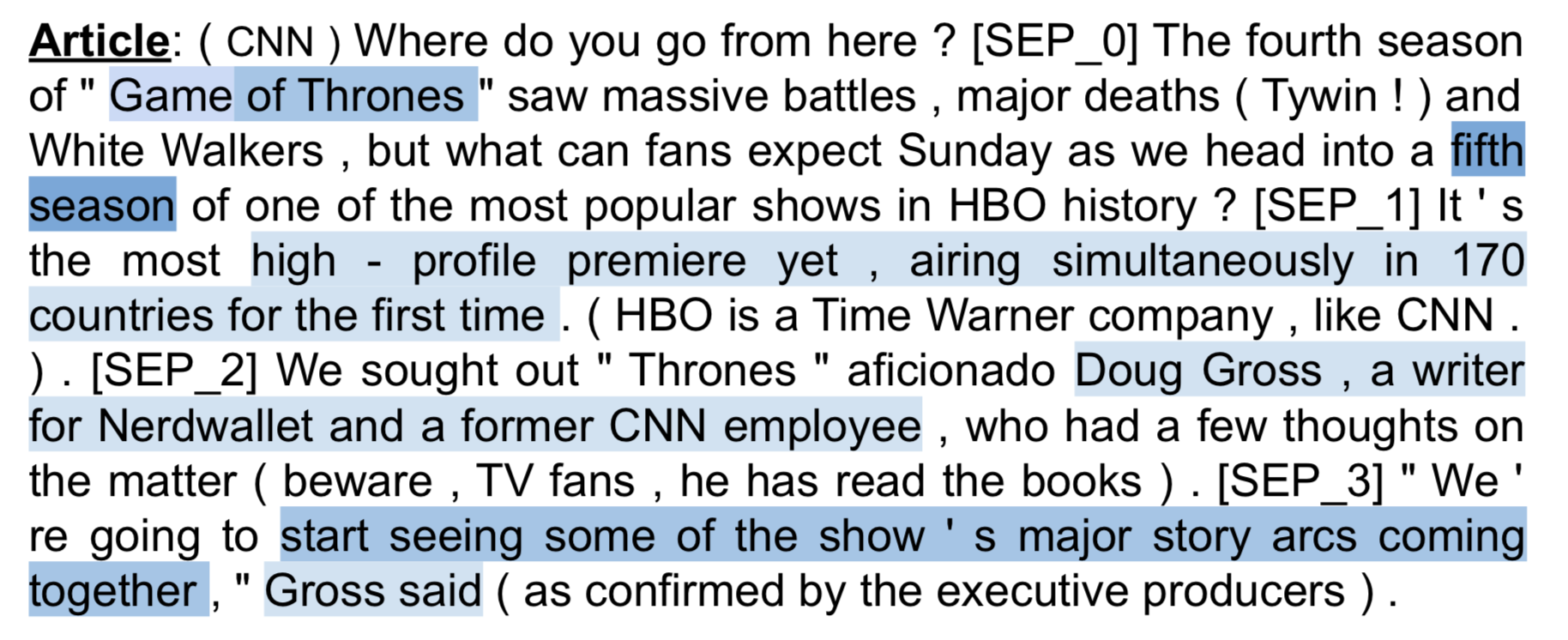}
\caption{Sample Article / Document}
\label{article}
\end{figure*}

\begin{figure*}[htp]
\centering
\includegraphics[width=15cm,height=10cm]{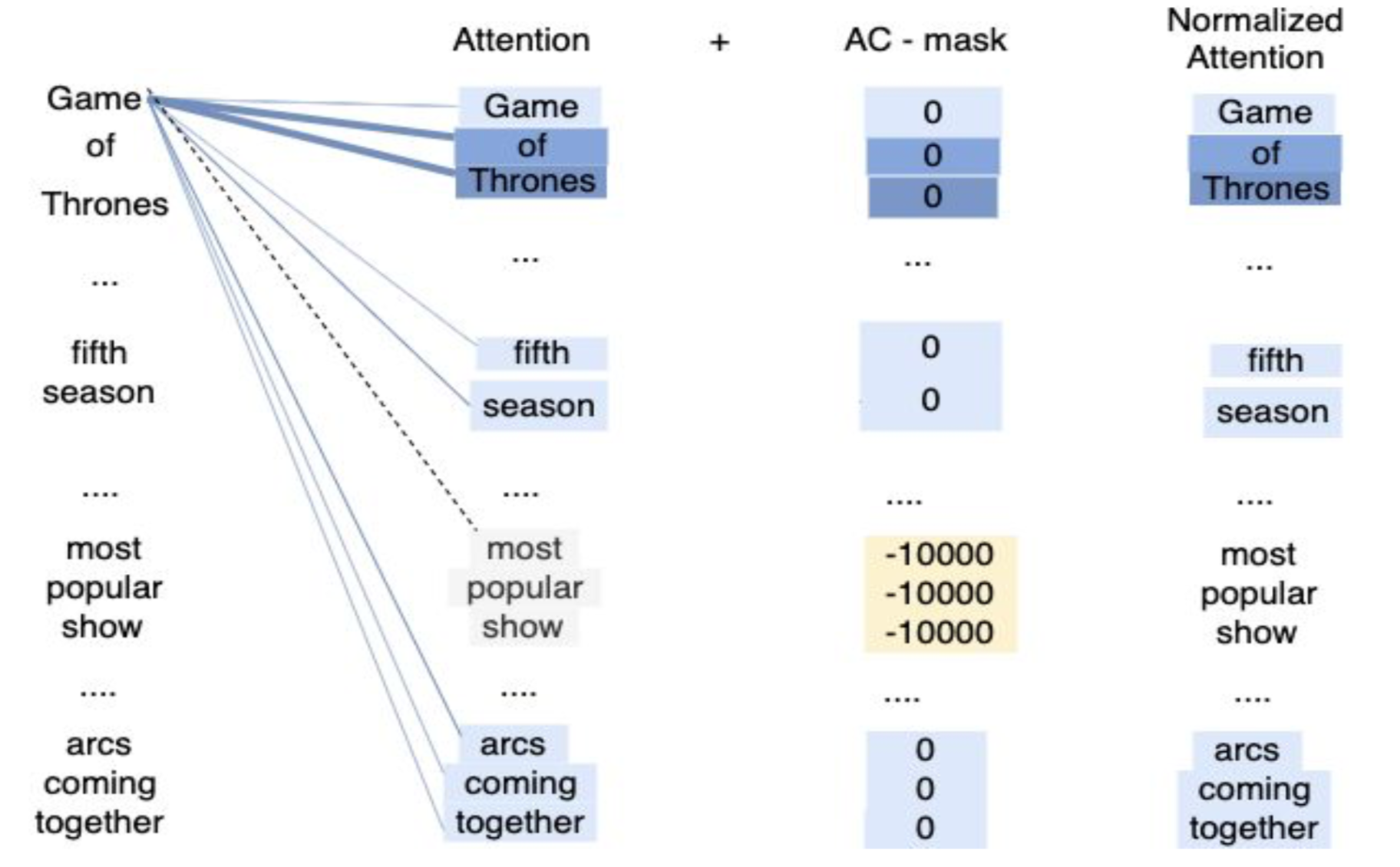}
\caption{Visualization of Attention Score Distribution with AC Learning
}
\label{vizattn}
\end{figure*}

Figure \ref{article} presents a snippet of a CNN article about the premier of season five of the HBO show Game of Thrones. The example illustrates how the model pays attention to specific words of the input in generating a multi-sentence summary. The A2C agents uses the attention map for one of the words (state) to select (sample) the attention mask from the learned policy distribution as shown in figure \ref{vizattn}.  \par

\begin{figure*}[htp]
\centering
\includegraphics[width=15cm,height=5cm]{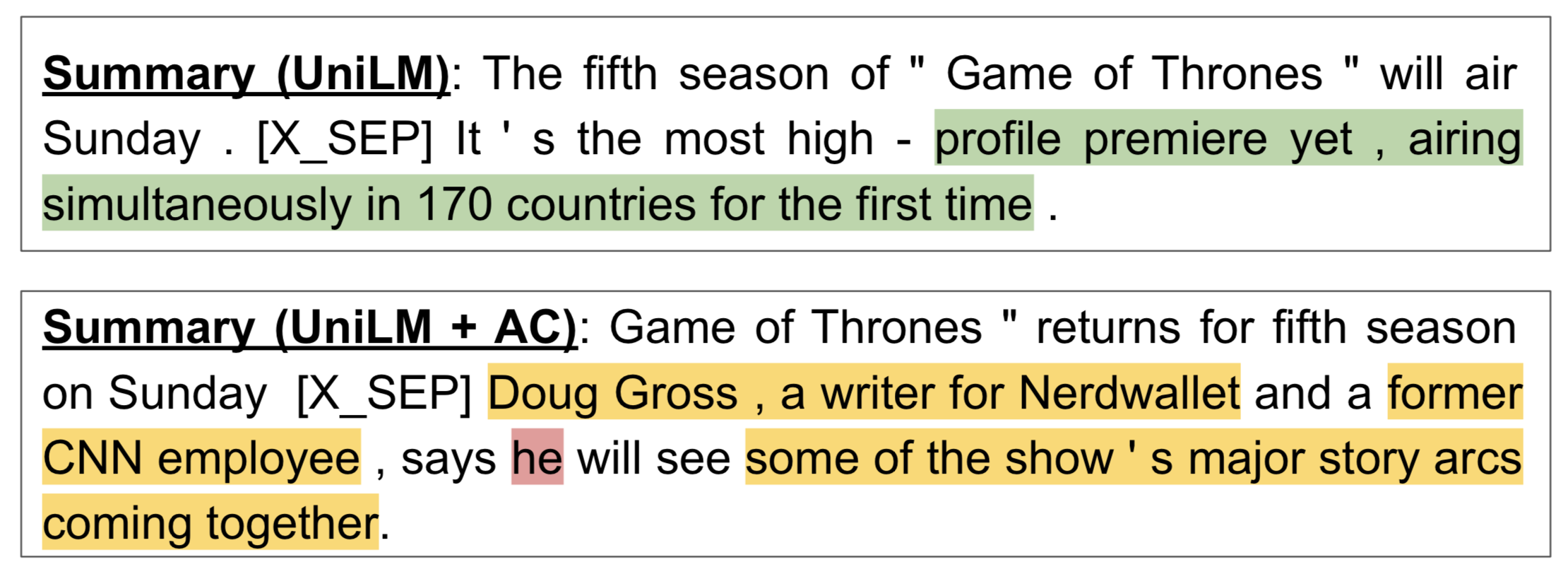}
\caption{Qualitative comparison between UniLM and DR.SAS - Game Of Thrones
}
\label{quality}
\end{figure*}

\begin{figure*}[htp]
\centering
\includegraphics[width=15cm,height=14cm]{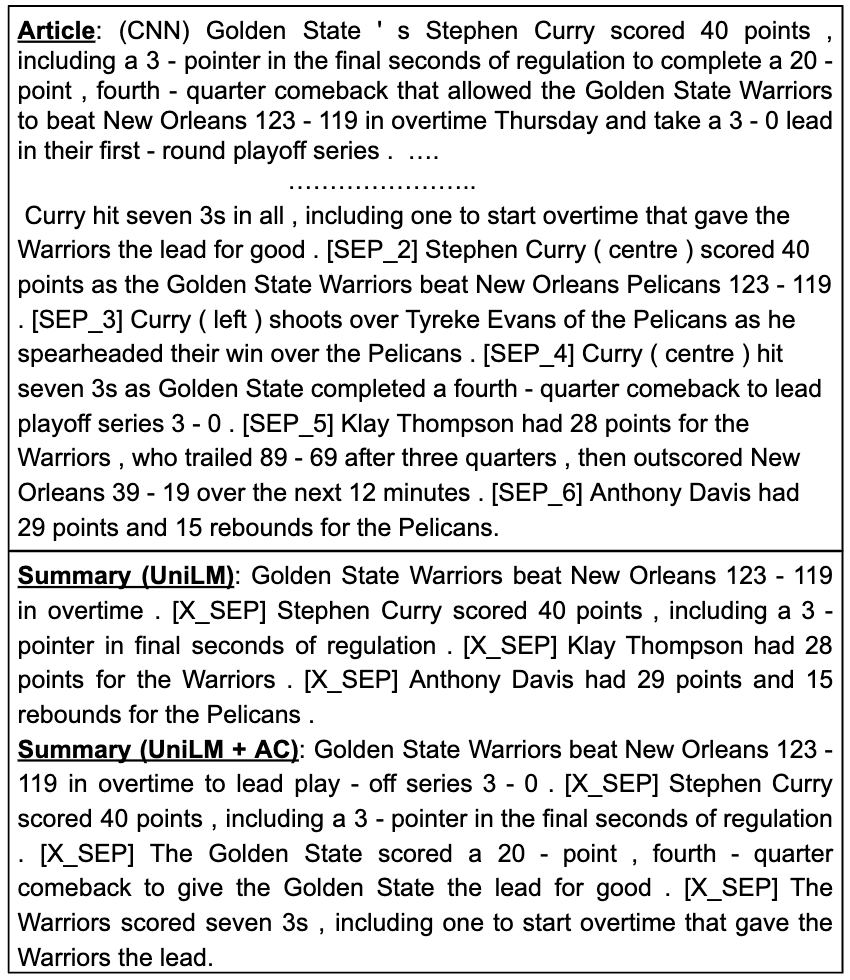}
\caption{Qualitative comparison between UniLM and DR.SAS - Golden State Warriors
}
\label{quality2}
\end{figure*}

 A closer look at the result summaries (figure \ref{quality} of UniLM and UniLM + A2C models show that the RL agents seems to generate masks that would increase emphasis on different parts of the input in the service of increasing the ROUGE score. That add some extractive power to the summary. The model chose to add focus on what 'Doug Gross' has to say about the fifth season premier of game of thrones. Another example in figure \ref{quality2} shows that UniLM + AC increases the ROUGE reward and may extract a bit more factual points but maybe not necessarily essential points. However, the desired level of extracted factual points may be domain and application specific. \par

\section{Conclusion and Future Work}
We present DR.SAS - a novel method that combines reinforcement learning with a summarization seq-to-seq model to learn dynamic self-attention masks. Our method allows us to generate coherent and factual summaries and address some limitations presented in literature where abstractive summarization models tend to be superfluous in terms of details. Our model achieves better ROUGE scores and also has more factual details in the summaries compared to baseline UniLM model. These learnings can be extended to other natural language tasks. As future work, we would like to experiment with layer specific self-attention to better understand the granularity of features most effective for summarization. 
Challenges:
\begin{enumerate}
\item Encoding embeddings or the meaning of tokens in state information would be a more precise way to take the methodology forward
\item A precise state space could lead to a very high number of parameters.
\item UniLM + AC is compute heavy and requires a lot of GPU resources to converge and generalize
The reinforced model utilizes ROUGE metric as a reward. ROUGE is biased towards more overlap between the reference summary and the model generated summary. That may add bias towards extracting more information from the original article (not necessarily essential points)
\end{enumerate}

\section{Contributions and Acknowledgements}
Both authors contributed equally to problem research and formulation, design and coding of the model architecture, development environment setup, data preparation and analysis, designing and conducting experiments, hyper-parameter tuning, evaluation and visual/error analyses and report writing. We are thankful to the CS221 Team for all the support and help throughout and to the folks at Google for the Cloud credits that made this work possible.

\end{document}